\documentclass{article}

\usepackage[nonatbib,final]{private_version_of_nips_2016}

\usepackage[utf8]{inputenc} 
\usepackage[T1]{fontenc}    
\usepackage{hyperref}       
\usepackage{url}            
\usepackage{booktabs}       
\usepackage{amsfonts}       
\usepackage{nicefrac}       
\usepackage{microtype}      
\usepackage{comment,alltt}

\title{Dataflow matrix machines as programmable, dynamically expandable, self-referential\\
generalized recurrent neural networks}

\author{
  Michael Bukatin\\
  HERE North America LLC\\
  Burlington, Massachusetts, USA\\
  \texttt{bukatin@cs.brandeis.edu} \\
  \AND
  Steve Matthews\\
  Department of Computer Science\\
  University of Warwick\\
  Coventry, UK\\
  \texttt{Steve.Matthews@warwick.ac.uk}\\
  \AND
  Andrey Radul\\
  Project Fluid\\
  Cambridge, Massachusetts, USA\\
  \texttt{aor021@gmail.com} \\
}

\begin{document}

\maketitle

\begin{abstract}
Dataflow matrix machines are a powerful generalization of recurrent neural networks.  
They work with multiple types of linear streams and multiple types of neurons, including
higher-order neurons which dynamically update the matrix describing weights and
topology of the network in question while the network is running. It seems that the
power of dataflow matrix machines is sufficient for them to be a convenient
general purpose programming platform. This paper explores a number of
useful programming idioms and constructions arising in this context.
\end{abstract}

\section{Introduction}

Dataflow matrix machines seem to be an attractive platform for general-purpose programming
and at the same time dataflow matrix machines are representable by matrices of numbers
in a manner similar to recurrent neural networks (RNNs).

This opens new possibilities for various interesting applications, such as learning general-purpose software
via methods proven effective for learning recurrent neural networks, 
and programming generalized recurrent neural networks within generalized recurrent neural networks themselves.

\subsection{Architecture of dataflow matrix machines}

A dataflow matrix machine (DMM) can work with various kinds of linear streams of data (the streams of data for
which a notion of linear combinations of several streams is well-defined).

We currently require to specify a finite number of {\em kinds} of linear streams, $D_1, \dots, D_N$. We also currently
require to specify a finite number of {\em types} of neurons, $C_1, \dots, C_M$. An input of a neuron
computes linear combinations of linear streams of a particular kind before passing the resulting stream to the neuron.
A type of neuron specifies how many inputs a neuron of this type has (zero or more),
what kind of streams each input accepts, how many outputs a neuron of this type has (one or more),
what kind of streams each output emits, and the built-in transformation of the collection of input streams into
the collection of output streams.

It is convenient to consider a countable-sized infinite network, where only a finite number of connections have non-zero
weights at any given moment of time, and only neurons which have some non-zero weights on some connections associated
with them are active. Hence only a finite subnetwork is active at any given time, while other neurons are considered to silently produce
streams of zero vectors (they don't need to be present in memory and don't spend processing resources).

So we consider a countable set of neurons of any type $C_i$, and hence we have a countable number of
inputs and a countable number of outputs of any given kind $D_j$, and a countable-sized matrix of
connection weights associated with each kind of streams $D_j$, with only a finite number of those weights being non-zero
at any given moment of time.

The topology of the network in question is determined by the sparsity structure of the matrices in question and can also
evolve with time. 

In this paper, one cannot combine streams of different kinds directly, the only way to perform transformations between streams of different kinds is via neurons
of appropriate types. We combine countable-sized matrices for all $D_j$ into the single countable-sized matrix $A$ by enforcing the
sparsity condition that the weights of connections between streams of two different kinds are always zero.

In this paper, we consider synchronous discrete time which is the most standard choice for RNNs.

\subsection{Diverse nature of linear streams}

In this paper, we consider dataflow matrix machines over real numbers.

There is a vector space $V$ associated with a given kind of linear streams, and a stream associates a representation of
a vector $v_t \in V$ with each moment of time $t$. We are interested in different kinds of vector spaces $V$: numbers, 
finite-dimensional vectors of various kinds (including spaces of finite matrices and tensors of fixed shape), and
infinite-dimensional vector spaces such the space of signed measures over some $X$.

Because of our interest in infinite-dimensional vector spaces we have to allow the situations where the representation of
$v_t$ in the stream is less than perfect. E.g. we might decide to represent signed measures over $X$ by probabilistic samplings
from those measures (with samples marked by the $+$ and $-$ flags to indicate whether they come from  the positive or
the negative part of the distribution in question). The linear combination of the streams will be represented by a stochastic sum of streams
in this case; namely, to pick the next sample from $\alpha_1*\mu_1 + \dots + \alpha_n*\mu_n$ we pick the sample associated with
$\mu_i$ with the probability $|\alpha_i |/ \sum_j |\alpha_j|$, and if $\alpha_i <0$ we toggle its $+/-$ flag.

\subsection{Programming the dataflow matrix machines}

It turns out that the many conventional programming constructions can now be expressed in this formalism.

The composition (``compound statement") is expressed via layers similar to ``deep networks".

The \texttt{if} and other conditional constructions correspond to redirecting the flow of information in the network
and can be achieved by using multiplicative masks on matrix columns. 

Multiplicative masks on the matrix itself serve as a convenient mechanism 
to express subnetworks.

Accumulators of values for a particular kind of streams are achieved by allowing the type of neurons expressing
identity transformation for this kind of streams, and by making sure that the weight of the connection between
the output and the input of the accumulator in question is 1.

\subsection{Self-referential facilities}

Consider a situation where a vector associated with a stream of the kind $D_1$
is a  matrix, $A$, with the same structure of indices for rows and columns as matrices describing the dataflow
matrix machines in question. Then we can think of
such vectors as descriptions of dataflow matrix machines, and we can think of neurons which take and produce streams of this kind as
transformers of dataflow matrix machines.

More specifically, let's consider the type $C_1$ of neurons to be the identity transformers of the streams of the kind $D_1$.
And let's fix a particular neuron \texttt{Self} of the type $C_1$, and fix the weight of the connection of its output to its
input to be 1 to make it an accumulator, and associate our dataflow matrix machine with the matrix accumulated on the output of
the neuron \texttt{Self}.

Then as long as we make sure to introduce types of neurons doing interesting transformations with streams of the kind $D_1$, we have facilities
for self-referential dynamic modification of a dataflow matrix machine while it runs. Like any accumulator, \texttt{Self}
treats contributions from other streams of the kind $D_1$ to which \texttt{Self}'s input is connected by non-zero weights
as additive modifications.

\subsection{Programming language situation}

We envision a situation where one has an interactive high-level language to describe and modify the dataflow matrix machine.
In this paper we start sketching fragments of one possible language of this kind.

The statements written in that language edit the matrix $A$ defining the dataflow matrix machine in question,
and neurons whose $D_1$-typed outputs have non-zero connections into \texttt{Self} are also capable of editing
that matrix.

\subsection{Powerful neurons}\label{powerful_neurons}

This approach envisions having a sizable collection of diverse neuron types.  The collection of kinds of streams and
types of neurons is called a {\em signature} of our dataflow matrix machine,
and our expectations for dataflow matrix machines to be a convenient
platform for general-purpose programming is based on the ability to add sufficiently expressive
types of neurons to the signature.

In particular, to participate in conditional constructions controlled by multiplicative masks over columns
(that is, over neuron outputs), a neuron should have a special input for the value of that
multiplicative mask. So a neuron would typically have a separate input computing the number associated with
the mask, and its associated transform would multiply the values of output streams of the neuron in question by that number.
An optimized implementation would be able to omit the associated computations when the mask is zero,
and to omit the multiplications when the mask is 1.

We envision having a sufficient collection of types of neurons capable of editing the matrix $A$ to easily
express the edits doable via the interactive high-level language mentioned above. We think about this
informally as ``self-referential completeness of the DMM signature relative to the language available to
describe and edit the DMMs". Note that it is usually easy to trigger a neuron which has a multiplicative
mask input for exactly one particular moment of time if necessary, e.g. by connecting an external input control 
into the multiplicative mask input of our neuron. An example of such a type of neurons is given in
Section~\ref{higher_order_neuron}.

\subsection{Bipartite graphs}

For the case when all neurons have one output, it is sometimes convenient to think about these networks as bipartite graphs.
One kind of nodes are neurons with finite fixed input arities. 

Another kind of nodes are {\em neuron inputs} which have variable
input arity. They work as countable linear combinations of the neuron outputs of the appropriate kind,
so formally speaking they have countable input arity,
but only finite number of coefficients are non-zero at any given moment of time. These linear combinations
are represented by rows of the matrix $A$.

A neuron input has one outgoing link, sending the linear combination of streams to the neuron itself. 
The situations where one neuron input serves multiple neurons correspond to constraints requiring different rows
of the matrix $A$ to be equal and are beyond the scope of this paper.

\section{Related work}

\subsection{Recurrent neural networks as a computational platform}

The ideas that recurrent neural networks can be used as a programming platform, and that a recurrent
neural network can modify other recurrent neural networks including itself are certainly not new.

Turing universality of some classes of recurrent neural networks  became known at least 20 years ago
and first schemas to compile conventional programming languages into recurrent neural networks
emerged in the same time frame (see~\cite{JNetoSiegelmannCosta} and references therein). 

That time period had also seen the first thought experiment in recurrent neural networks involving
``self-referential" weight matrix where the network had used its own weights as additional input data and could,
at least in principle, learn new algorithms for modifying its own weights~\cite{JSchmidhuber}.

The first successful instance of using a recurrent neural network to learn a well working learning
algorithm for training some recurrent networks appeared in~\cite{SHochreiterYoungerConwell}.

However, none of these became a part of the daily routine of the field. One would be hard-pressed to find recent
instances of using RNNs to either manually create general-purpose software or to learn the algorithms of RNN synthesis.
In fact, a recent influential post~\cite{AKarpathy} emphasizes that training recurrent networks should be considered
optimization over a space of programs, but at the same time cautions  not to ``read too much into" the theoretical
results establishing Turing-completeness of RNNs.

On the positive side, there is quite a bit of interest recently in using recurrent neural networks and
related machines to learn general-purpose algorithms automatically (see e.g.~\cite{SReeddeFreitas} and references
therein). The question seems to be, how much would one need to modify the traditional RNN architecture
and learning methods to make general-purpose program learning feasible for daily use.

We think that the difficulties are to a large extent related to the focus on one-dimensional streams
(streams of numbers), to configurations with fixed number of neurons, and to per-element
matrix modification schemas. Combination of these factors usually leads to awkward encodings,
which make the network behavior overly sensitive to very small changes.

For example, one might note that Turing universality requires potentially infinite memory. In a typical approach
to Turing universality of recurrent neural nets one expands the precision of the involved real numbers
in an unbounded fashion, so that one could  in effect use the binary expansion of a real number as a tape
for a Turing machine~\cite{HSiegelmann}.

Another way would be to expand the size of the neural net in question in an unbounded fashion.
Since we already have a countable-sized matrix, this approach seems more natural for our
architecture.
This approach seems also to be more consistent with the traditional uses of neural nets,
which involve the standard built-in machine representation of real numbers (or, sometimes,
the compressed versions of even that, for better performance), and this is the approach
we use in the present paper.

We also tend to modify the network matrix not on per-element basis, but by functional blocks
in the present paper, and we expect this to also help us to avoid awkward encodings and delicate
coordination problems, where a desired modification goes through only for some of the elements
expected to be involved in this modification,
creating a configuration with uncertain properties.

\subsection{Dataflow programming with linear streams}

Dataflow programming languages such as LabVIEW and Pure Data (which are oriented mostly towards work with streams of continuous data)
found some degree of general programming use within their application domains~\cite{WJohnstonHannaMillar,AFarnell}.
However, those platforms are not purely continuous, and it's not clear how to achieve matrix parametrization of large spaces
of programs for those platforms.

The discipline of dataflow programming purely in terms of diverse variety of linear streams was considered in~\cite{BukatinMatthewsLinear},
where it was also noted that adopting a discipline of bipartite graphs with one class of nodes being associated with general transformations
and another class of nodes being associated with linear transformations allows to represent large classes of dataflow programs
over linear streams by matrices of real numbers and to modify programs by continuous change of those numbers, thus achieving
continuous program transformations.

However, it was not demonstrated in~\cite{BukatinMatthewsLinear} that its approach would lead to
a viable general-purpose programming platform, nor was it mentioned in~\cite{BukatinMatthewsLinear} that the architecture developed there
was a generalization of recurrent neural networks.

In this paper, we observe that the architecture in the spirit of~\cite{BukatinMatthewsLinear} generalizes recurrent neural networks
and make progress towards developing this architecture into a viable general-purpose programming platform.
A particularly novel contribution
is self-referential modification of matrices controlling the network behavior not on per-element basis, but by additive functional
blocks. The ability to do so is enabled by going beyond one-dimensional streams of numbers and allowing single neurons to
work with streams of countably-sized matrices containing finite number of non-zero elements.

\section{A more detailed description and programming of DMMs}

\subsection{Neuron types}

Consider kinds of streams $D_1, \dots, D_N$ equipped with the ability to take linear combinations
of streams of the same kind with finite number of non-zero coefficients.

Define a type of neurons $C_i$ as follows. Associate with $C_i$ the non-negative integer number of inputs $\#I_i$ (neurons which
are to unconditionally function as inputs for the network have zero inputs)
and the positive integer number of outputs $\#O_i$. With every input or output $k$ of $C_i$ associate a kind of streams $D_{j_k}$.
Also associate with $C_i$ a transform $F_i$ taking as inputs $\#I_i$ streams  of appropriate kinds of length $t-1$ and producing as outputs $\#O_i$ streams of appropriate kinds
of length $t$ for integer time $t>0$. Require
the obvious prefix condition that when $F_i$ is applied to streams of length $t$, the first $t$ elements of the output streams of length $t+1$ are the elements
which $F_i$ produces when applied to the prefixes of the input streams of length $t-1$.
The most typical situation is when for $t>1$ the $t$'s elements of the output streams are produced solely
on the basis of elements number $t-1$ of the input streams, but our definition also allows neurons to
accumulate unlimited history, if necessary.

The resulting $((D_1, \dots, D_N), (C_1, \dots C_M))$ is called the {\em signature} of our DMM.

We are going to define some countable collections and we find it convenient to index them by strings rather than
by numbers. Formally speaking, strings are equivalent to numbers as a countable set
of indices, since the alphabet for strings can be thought of as digits in the appropriately
selected system of numbers. But the approach based on strings does
de-emphasize the order on indices coming from those numbers and encourages
to use indices for meaningful description of functionality without excessively
cumbersome G\"{o}del-style numbering.

We use a base alphabet $\Sigma$ for those strings and additional special characters we always assume to be
outside of $\Sigma$. In this paper those special characters are the space and \texttt{\#;:,=} characters.

\subsection{Describing the DMM elements in a programming language}

Keywords in our DMM-oriented language start with the special character \texttt{\#} and are followed by the space.

The kinds of streams are named \texttt{\#kind <name1>, ..., \#kind <name N>}. 
The user must supply the implementation of the respective streams
and their linear combinations in the underlying conventional language.

The types of neurons are named \texttt{\#celltype <typename 1>, ..., \#celltype <typename M>}.

The types of neurons are specified by stating stream kinds and giving the names to their inputs (if any) and their outputs:
\begin{verbatim}
    #newcelltype <typename> 
        #input  <kindname>:<fieldname> ... #input  <kindname>:<fieldname> 
        #output <kindname>:<fieldname> ... #output <kindname>:<fieldname>;
\end{verbatim}

The user must supply the implementation of the stream transform associated with
each neuron type in the underlying conventional language.

The neurons are named using special character \texttt{:} as follows: \texttt{\#neuron <typename>:<cellname>}, where
\texttt{<cellname>} is any string in the alphabet $\Sigma$. We assume that all such neurons silently exist,
and can be activated by making some of their connections non-zero.

The matrix element $a_{ij}$ in question describing a connection from an output of a neuron (cell 1) to an input of a neuron (cell 2) can be written as
\begin{verbatim}
    #weight <typename 2>:<cellname 2>:<inputfieldname>
            <typename 1>:<cellname 1>:<outputfieldname>
\end{verbatim}

This weight can only be non-zero when the kinds of streams associated with the fields
 $i$ = \texttt{<typename 2>:<cellname 2>:<inputfieldname>} and $j$ = \texttt{<typename 1>:<cellname 1>:<outputfieldname>}
coincide. Otherwise, the weight is forced to be zero.

\subsection{Dynamic modifications of a working DMM}

To describe a work cycle of a dataflow matrix machine, it is convenient to have the metaphor of
``two-stroke engine" in mind. In terms of graphical depiction of the DMM, think about a horizontal layer of all neurons
with outputs being above that layer and inputs being below that layer.
 Then on the ``up movement", neurons compute their outputs based on their
inputs using transforms associated with their types. On the ``down movement", {\em neuron inputs}
compute themselves as linear combinations of all neuron outputs using the rows of matrix $A$.

Consider a neuron with the identity transform for some kind of streams.
It is important to note that even though the transform is the identity, one can't just
use the single shared variable to store the latest values of the input and output
streams of that neuron. On the ``down movement", all neuron inputs including the input for the
neuron in question are recomputed as linear combinations of all outputs; it is important for
the corresponding output of the neuron in question not to be overwritten
during the ``down movement" because it might be used in computing a variety of linear combinations.
Only during the ``up movement" does the identity transform get executed changing the neuron output.

In particular, this is applicable to the neuron \texttt{Self} controlling the matrix $A$ itself. The output of
\texttt{Self} holds the current value of the matrix $A$. The rows of $A$ are used to compute all the inputs of the
network on the ``down movement", including the input of \texttt{Self} which accumulates the value
of the matrix $A' = A +$ contributions from other neurons. 
Then on the ``up movement", the identity transform updates the $A$ to become $A'$. Making this process
as efficient as possible is a separate topic.

\subsection{Initialization and expandable memory}

Neurons with all zero incoming and outgoing weights (that is, with all rows of $A$ corresponding to inputs
of the neuron in question and with all columns of $A$ corresponding to outputs of the
neuron in question being zero) are only present in the countable address space, but not in memory.
Their outputs are considered to be zero vectors, and they don't participate in computations.

A neuron gets into memory and becomes active when one of its associated weights becomes
non-zero. Since this can only happen by a matrix update, the first computation which happens
after that is ``down movement", so first the inputs of the neuron in question are updated,
only then the neuron itself works.

There is an option to ``garbage collect" a neuron from memory, if all its associated rows
and columns are set to zero.

\subsection{Programming}

There is a variety of approaches to neuron names within the DMM, ranging from completely semantically meaningful names to
autogenerated tokens, and there are various trade-offs associated with this choice. In this paper
we explore how far we can go using autogenerated tokens and avoiding the reliance on semantically
meaningful names. 

At the same time, in our DMM-oriented language we need identifiers to denote various parts of
the DMM in question, e.g. a neuron, a particular input or output of a neuron, a subgraph, etc.

To avoid ambiguity we always use the word {\em name} for indices of columns and rows in our matrix
and similar purposes and we always use the word {\em identifier} for variables in our DMM-oriented
language.

All neurons and links (countably infinite sets) are presumed to exist, and the burden of not spending
space and processors on all but the finite number of those is on the implementation.

Therefore a statement 
\begin{verbatim}
    #neuron <typename>:<IdNeuron> 
        <fieldname>:<IdOutput1>, ..., <fieldname>:<IdOutputO> =
        #transformof <fieldname>:<IdInput1>, ..., <fieldname>:<IdInputI>; 
\end{verbatim}
simply introduces an identifier \texttt{<IdNeuron>} to denote a neuron with a new autogenerated \texttt{<cellname>} and zero associated weights,  
and it also introduces identifiers \texttt{<IdOutput1>, ..., <IdOutputO>, <IdInput1>, ..., <IdInputI>} for the input and output streams 
of this neuron in our language, but it does not change the underlying network.

Similarly,
\begin{verbatim}
    #stream <kindname>:<IdInputStream> = #neuroninput <IdNeuron>.<fieldname>;
\end{verbatim}
introduces an identifier for a particular
input stream of the neuron denoted by \texttt{<IdNeuron>}.

But no functional changes can be obtained in this fashion. The only way to make the network to change its behavior is
to update the matrix weights. We always do it respecting our accumulator metaphor, for example to update
the way a particular input to a neuron is computed from a variety of neuron outputs with the same kind of streams we can write
\begin{alltt}
    #updateweights <IdInputStream> +=
        \( \alpha\sb{1} \)* <IdOutputStream\(\sb{1}\)> + \dots +\( \alpha\sb{k} \)* <IdOutputStream\(\sb{k}\)>;        (1)
\end{alltt}

Or one can add a linear combination of rows corresponding to other input streams of the same kind:

\begin{alltt}
    #updateweights <IdInputStream> +=
        \( \beta\sb{1} \)* <IdInputStream\(\sb{1}\)> + \dots +\( \beta\sb{k} \)* <IdInputStream\(\sb{k}\)>;          (2)
\end{alltt}

This provides one of the mechanisms to set a row to zero:

\begin{alltt}
    #updateweights <IdInputStream> += (-1) * <IdInputStream>;          (3)
\end{alltt}

\subsection{Streams of rows and streams of columns}

In addition to streams of matrices capable of controlling the DMM in question it is convenient to have streams of rows and streams of columns,
so these are two more kinds of streams to have. The values are used directly as rows and columns of the matrices controlling the DMM and
also as multiplicative masks (that is, coefficients of linear combinations; rows are used as ``column masks", and columns are used as ``row masks").

The typical values have only finite number of non-zero elements, but some finitely describable
infinite vectors are quite helpful in this context, e.g. infinite rows with all fields corresponding
to outputs having a particular kind of streams equal to 1 are useful both as trivial column masks to be
applied to finite row vectors and
as fake input rows (in the example \texttt{(1)} of the \texttt{\#updateweights} operation
a finite column mask (vector $\alpha$) is applied to a fake input row of this kind). 

The discipline of only
allowing non-zero elements for one particular kind of streams in each of our rows and columns might be
preferable (otherwise one needs to worry about not breaking the prohibition of creating
non-zero weight elements between streams of different kinds).

The general form of the \texttt{\#updateweights} statement covering the examples above and more is

\begin{alltt}
    #updateweights <FiniteRowMask 1> += <ColumnMask> * <FiniteRowMask 2>;
\end{alltt}

In all the \texttt{\#updateweights}  examples above, \texttt{<FiniteRowMask 1>} has 1 in one place and zeros in the other places, 
but any finite row mask is allowed and
coefficients can be different from 1 and 0 (they are used as multipliers for the right-hand side before the
\texttt{+=} is applied for the given row).

In the example \texttt{(2)} of \texttt{\#updateweights} the vector of $\beta$'s at the \texttt{<IdInputStream$_i$>} positions works as a \texttt{<FiniteRowMask 2>}, and the
example \texttt{(3)} is a particular instance of the example \texttt{(2)}. The \texttt{<ColumnMask>} is a trivial mask with infinite number of 1 fields described above in both cases.

The \texttt{\#updateweights} operation is oriented towards working with rows, but a similar operation oriented towards working with
columns can be introduced if necessary.

Take \texttt{<ColumnMask>} to be vector $\alpha$, \texttt{<FiniteRowMask 2>} to be vector $\beta$, and the left-hand-side 
\texttt{<FiniteRowMask 1>} to be vector $\gamma$. Then on the element-wise level the \texttt{\#updateweights} operation acts as
$a_{ij} := a_{ij} + \gamma_i * \alpha_j * \sum_k \beta_k a_{kj}$.

\subsection{Editing the DMM matrix with higher-order neurons}\label{higher_order_neuron}

The paradigm formulated in Section~\ref{powerful_neurons} requires us to have a type of higher-order neurons to perform
the \texttt{\#updateweights} operation. The neurons of this type have to be fairly complex, they need to have at least 5 different inputs.
One input is to be of the kind corresponding to the stream of matrices controlling the DMM. Most typically, the output of \texttt{Self}
will be connected to this input with the weight of 1. Two more inputs are needed for
\texttt{<FiniteRowMask 1>} and \texttt{<FiniteRowMask 2>}, and one more input for \texttt{<ColumnMask>}.
Finally, at least one more input is needed to control the firing of these neurons, for example a scalar
multiplicative mask as discussed in Section~\ref{powerful_neurons}.

These neurons would have one output of the kind corresponding to the stream of matrices controlling the DMM.
This output would normally emit the result of the application of the \texttt{<FiniteRowMask 1>}
to the result of the right-hand side of the \texttt{\#updateweights} operation (the result of the application of the \texttt{<ColumnMask>} to the result of the application of
the  \texttt{<FiniteRowMask 2>} to the matrix itself).
Most typically, this output would be connected to the input of \texttt{Self} with the weight of 1.

\section{Conclusion and future work}

Dataflow matrix machines and recurrent neural networks represent a continuous programming architecture, that is,
an architecture oriented towards working with the streams of continuous data, with the programs being of
the continuous nature themselves. Discrete programming architectures
support a variety of programming styles: imperative, objective-oriented, functional, logical, dataflow, etc.
Similarly, a variety of programming styles should be possible for DMMs and RNNs.

For example, it is relatively easy to support an imperative programming style based on representing
linked data structures in the body of a DMM and having neurons traversing those structures, based
on the machinery introduced in the present paper.

It is also relatively easy to support a programming style based on creating deep copies of subgraphs
while preserving their incoming (and, optionally, outgoing) connections.
One can easily create fairly intricate patterns by using this operation
in a nested way.\footnote{{\bf June 2018 note:} See Section 4 of \url{https://arxiv.org/abs/1606.09470} for more details on deep copies of subgraphs. 
See Appendix D of \url{https://arxiv.org/abs/1610.00831} for the first implementation of a self-modifying DMM in the style of Section 3.3 of the present paper 
(see Appendix B of \url{https://arxiv.org/abs/1706.00648} for a more polished implementation and presentation of that).

Reference paper on DMM research in 2016-2017 is
\url{https://arxiv.org/abs/1712.07447}}

We hope that more styles of programming in this architecture will emerge.

One of the hopes of this approach is that it eventually results in better ways for
program learning. Currently, a typical result of program learning is a program which
functions, but is almost impossible for a human to read and understand.
At the same time, recurrent neural networks have good track record of learning patterns,
and of generating visually readable patterns back (see e.g.~\cite{AKarpathy}).

One would hope that if one starts with a corpus of programs in a DMM-oriented
language, one could then find a way to train a DMM to reproduce both syntactic patterns and function,
resulting in programs which not only work, but can also be read and understood by people.

\end{document}